\documentclass{article}
\usepackage{iclr2027_conference,times}

\usepackage{amsmath,amsfonts,bm}

\def\eqref#1{equation~\ref{#1}}

\def\1{\bm{1}}

\DeclareMathAlphabet{\mathsfit}{\encodingdefault}{\sfdefault}{m}{sl}
\SetMathAlphabet{\mathsfit}{bold}{\encodingdefault}{\sfdefault}{bx}{n}

\usepackage{hyperref}
\hypersetup{hidelinks}
\usepackage{url}
\usepackage{booktabs}
\usepackage{graphicx}
\usepackage{wrapfig}
\usepackage{placeins}
\usepackage{float}
\usepackage[font=small]{caption}

\makeatletter
\renewcommand{\paragraph}{\@startsection{paragraph}{4}{\z@}%
  {2.0ex \@plus 0.6ex \@minus 0.2ex}{-0.8em}{\normalfont\normalsize\bfseries}}
\makeatother
\AtBeginDocument{%
  \setlength{\abovedisplayskip}{6pt plus 2pt minus 3pt}%
  \setlength{\belowdisplayskip}{6pt plus 2pt minus 3pt}%
  \setlength{\abovedisplayshortskip}{3pt plus 1pt}%
  \setlength{\belowdisplayshortskip}{3pt plus 1pt}%
}
\usepackage{amsmath}
\usepackage{amssymb}
\usepackage{xcolor}

\newcommand{\best}[1]{\textbf{\boldmath #1}}

\newcommand{\ours}{JEPA-\textit{x}}
\newcommand{\visual}{\textsc{Visual}}
\newcommand{\regress}{\textsc{Regress}}
\newcommand{\distill}{\textsc{Distill}}
\newcommand{\shuffle}{\textsc{Shuffle}}

\newcommand{\yes}{\checkmark}
\newcommand{\no}{\textendash}
\newcommand{\perm}{$\times$}

\title{\ours{}: Cross-Predictive Physics Grounding for Forecastable Latent Dynamics}

\author{
Kehan Wen\,\textsuperscript{1}, Ziming Li\,\textsuperscript{1}, Siyuan Luo\,\textsuperscript{1}, Fan Shi\,\textsuperscript{1}\\
\normalfont\textsuperscript{1}Department of Electrical and Computer Engineering, NUS
}

\newif\ificlrpreprint
\def\iclrpreprint{\iclrpreprinttrue\iclrfinalcopy}
\newcommand{\iclrpreprintnote}{Preprint.}

\iclrpreprint

\begin{document}

\maketitle
\ificlrpreprint\lhead{\iclrpreprintnote}\fi
\suppressfloats[t]

\begin{abstract}
Latent world models plan by predicting how candidate actions advance learned latent dynamics. In self-predictive models, however, the encoder and predictor are optimized jointly and can co-adapt to latent transitions that are easy to predict but weakly constrained by the physical evolution of the scene. We introduce the cross-predictive JEPA (\ours{}), which grounds latent dynamics in privileged physical trajectories. \ours{} treats visual observations and physical states as corresponding views of the same action-conditioned trajectory, advances both through a shared predictor, and matches each prediction to the future representations of both modalities. This encourages the action-conditioned predictor to learn a common transition rule across the two views.  Privileged physical state is used only during training, leaving a visual-only model at deployment. Empirical results show that \ours{} reduces the rollout drift of a newly fitted predictor from $0.361$ to $0.104$ and increases mean control success from $53.6\%$ to $78.2\%$ on a multi-task suite spanning six evaluation subfamilies. We additionally show that direct physical-state regression improves decodability without improving forecastability or control, indicating that the benefit comes from shaping latent dynamics rather than merely encoding physical variables.
\end{abstract}

\section{Introduction}
\label{sec:introduction}

Planning with a latent world model casts control as search in representation space \citep{ha2018world,hafner2019learning,hafner2023mastering,hansen2024tdmpc2}: the agent encodes the current scene, predicts the outcomes of candidate actions, and selects the actions whose predicted outcomes most closely approach the goal. This requires the world model to learn representations with reliable action-conditioned dynamics. In self-predictive models \citep{lecun2022path,assran2023self,zhou2024dinowm,sobal2025pldm}, however, the encoder that defines the prediction target is optimized jointly with the predictor. The encoder and predictor can therefore co-adapt to a latent transition structure that is easy for the co-trained predictor to model. Although this produces temporal consistency within the learned latent space, it does not directly constrain the transitions to follow the physical evolution of the scene and potentially limits the model's generalization capacity on unseen scenerios.

Privileged physical states have been widely used to shape learned representations and improve downstream behavior through state regression, distillation, cross-modal alignment, and training-time asymmetric critics~\citep{chen2020learning,gupta2016cross,tian2020contrastive,kumar2021rma,pinto2018asymmetric}. Despite their differences, these approaches primarily use privileged information to supervise what a representation encodes or how it is used downstream. Our empirical results, however, show that making physical state decodable from visual latents does not necessarily make their action-conditioned evolution easier to predict. This gap reflects a distinction among three properties: decodability asks whether physical variables can be recovered from an individual latent state; forecastability asks whether the representation supports reliable action-conditioned prediction beyond the predictor with which it was trained; and control utility depends on whether predicted outcomes preserve the distinctions needed to rank candidate actions. Improving one property does not necessarily improve the others. We therefore treat visual observations and physical states as corresponding views of the same action-conditioned trajectory, allowing the physical future to constrain how visual representations evolve under action rather than only what physical variables an individual latent state encodes. We then examine whether this trajectory-level constraint produces more forecastable visual dynamics and whether those dynamics improve control.

We introduce the \emph{cross-predictive JEPA} (\ours{}) to apply this trajectory-level constraint. The same action-conditioned predictor advances latent histories from either modality, and each prediction is matched to future representations in both modalities. The within-modal terms preserve self-prediction, while the cross-modal terms couple the predicted transitions across modalities. Sharing the predictor encourages the two encoders to expose a common action-conditioned transition rule rather than only align individual states. The physical branch is used only during training. At deployment, it is removed, leaving the same visual encoder--predictor architecture and planner as the visual-only baseline.

To test whether the improvement lies in the representation rather than its co-trained predictor, we freeze each visual encoder and train the same predictor from scratch. \ours{} reduces rollout drift from $0.361$ to $0.104$ under this evaluation. A single model trained across multiple object--interaction pairs also raises mean control success from $53.6\%$ to $78.2\%$, with improvements across all six evaluation subfamilies. In contrast, regression to the same physical state target improves decodability but leaves forecastability and control near the visual-only baseline. Ablations further show that correct visual--physical pairing is important for control, while direct cross-modal prediction improves forecastability beyond frame-level alignment.

In summary, our contributions are:
\begin{itemize}

\item \textbf{Cross-predictive JEPA.}
We introduce a vision--physics prediction objective that uses a shared predictor to couple action-conditioned transitions across modalities, with privileged state required only during training.

\item \textbf{A unified privileged-state interface.}
We design a shared rigid-body representation that allows one physical encoder to operate across diverse manipulation scenes without task-specific labels.

\item \textbf{More forecastable representations and better control.}
By evaluating frozen representations with newly trained predictors, we show that \ours{} improves visual-dynamics forecastability on the representation level. These gains lead to higher control success rate across the multi-task suite.

\end{itemize}

\section{Related Work}
\label{sec:related}

\paragraph{Predictive latent world models.}
World models differ in the structure that their learned representations preserve. Reconstruction-based approaches learn dynamics through pixels or generative latent variables \citep{ha2018world,hafner2019learning,hafner2020dream,hafner2023mastering}, whereas value-centric methods shape representations around reward and control objectives \citep{schrittwieser2020mastering,hansen2024tdmpc2}. Representation-predictive methods instead model future features directly: DINO-WM plans in pretrained visual features \citep{zhou2024dinowm}, PLDM learns latent dynamics from reward-free offline data \citep{sobal2025pldm}, and LeWM jointly learns visual representations and predictive dynamics \citep{maes2026lewm}. More broadly, joint-embedding objectives learn representations by predicting feature-space targets rather than reconstructing observations
\citep{lecun2022path,assran2023self,bardes2023mcjepa,bardes2024vjepa}.
Our work builds on this predictive paradigm and asks how corresponding physical trajectories can constrain the transition structure learned by a jointly optimized visual encoder and predictor.

\paragraph{Training-time privileged supervision.}
Learning with privileged information uses signals available during training but unavailable at deployment \citep{vapnik2015learning}. Prior work transfers such information through distillation, cross-modal alignment, state regression, privileged sensing, or asymmetric actor--critic training \citep{chen2020learning,lee2020learning,gupta2016cross,tian2020contrastive,kumar2021rma,pinto2018asymmetric}. These approaches establish several ways to exploit additional physical information during training, but our multi-task setting introduces a further requirement: the privileged representation must retain a consistent meaning across heterogeneous object--interaction configurations. We therefore represent scenes through a shared rigid-body schema based on object geometry, extent, pose, and effector state, allowing a single physical encoder to provide the privileged stream across all configurations without task-specific labels.

\paragraph{Physical grounding of predictive dynamics.}
Several recent world-model approaches bring privileged physical information closer to the learned dynamics. TWIST transfers state-based dynamics to an image-based student \citep{yamada2024twist}, Scaffolder uses privileged sensors during policy learning \citep{hu2024privileged}, PIGDreamer aligns world-model representations with privileged information \citep{huang2025pigdreamer}, and Pri4R introduces privileged 4D prediction during vision--language--action training \citep{kim2026pri4r}. Closest to our setting, the concurrent Phys-JEPA imposes physical consistency on latent states and transitions for multivariate time-series forecasting \citep{nie2026physjepa}, while PhyLatent adds training-only physical grounding and future-alignment objectives to the LeWM backbone \citep{zeng2026phylatent}. \ours{} instead treats privileged state as a second online predictive view: histories from either modality predict the corresponding future representations in both modalities, directly coupling visual--physical pairing with action-conditioned evolution. Unlike one-way distillation toward a fixed physical target, both representations remain online, and the privileged branch is removed entirely at deployment. We further study this coupling in the multi-task setting, where one model and one unified privileged-state interface span many object--interaction configurations.

\section{Preliminaries}
\label{sec:preliminaries}

\paragraph{Joint-embedding predictive architectures.} ~\citep{lecun2022path}
Joint-embedding predictive architectures (JEPAs) learn representations by predicting a target embedding from an encoded context rather than reconstructing the target observation. In temporal settings, the context is a latent history and the prediction can be conditioned on actions. The encoder and predictor are trained jointly, with an additional mechanism used to prevent representational collapse.

\paragraph{LeWM.} ~\citep{maes2026lewm}
LeWM applies this framework to action-conditioned visual prediction. A visual encoder maps each observation to $z_t^o=f_\theta(o_t)$, and a predictor estimates the future latent from a visual history $Z_t^o$ and an action chunk $a_t$. Its training objective is
\begin{equation}
\mathcal{L}_{\mathrm{LeWM}} = 
\left|
g_\psi(Z_t^o,a_t)-z_{t+1}^o
\right|_2^2
+
\beta\Omega(z^o),
\label{eq:lewm}
\end{equation}
where $\Omega$ is the SIGReg isotropy regularizer~\citep{balestriero2025lejepa}. For planning, candidate action sequences are rolled forward in the learned latent dynamics and scored against an encoded goal. We use LeWM as the visual-only baseline and retain its encoder, predictor, and planning procedure.

\section{Cross-Predictive JEPA}
\label{sec:method}

\begin{figure}[ht]
\centering
\includegraphics[width=\linewidth]{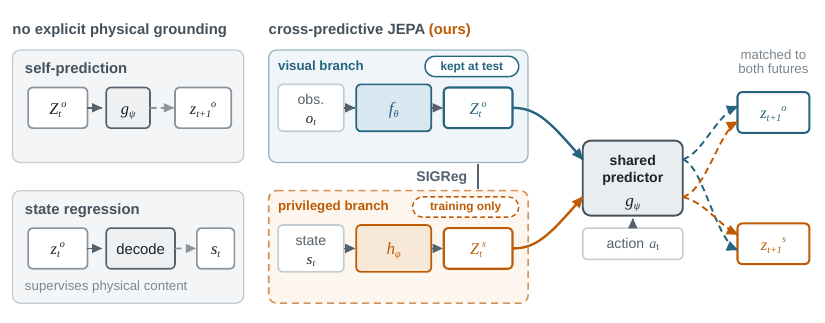}
\caption{\textbf{Cross-predictive physical grounding constrains representation geometry and predictive dynamics.}
Visual self-prediction jointly learns a visual encoder and predictor, while state regression makes physical variables decodable from individual visual latents. Neither directly couples a predicted visual transition to the corresponding physical future. \ours{} instead treats visual and physical streams as paired views of the same action-conditioned trajectory and matches predictions from either history to future representations in both modalities. The physical branch is removed after training, leaving a visual-only model at deployment.}
\label{fig:method}
\vspace{-0.2cm}
\end{figure}

\ours{} extends LeWM with a physical encoder and four within- and cross-modal prediction terms. The physical branch is used only during training and removed at deployment. Figure~\ref{fig:method} provides an overview.

\paragraph{Core objective.}
\label{sec:fourway}

For each visual trajectory, the simulator provides a synchronized sequence of privileged physical states. A physical encoder $h_\phi$ maps each state $s_t$ to a latent representation

$$
z_t^s=h_\phi(s_t),
$$

with the same dimensionality as the visual latent $z_t^o$. The privileged state describes only the instantaneous scene configuration through the unified interface introduced below. It excludes velocities, contact forces, wrenches, and goal-relative quantities. Motion must therefore still be inferred from the state history and the applied actions.

Let $Z_t^m$ denote a latent history from modality $m\in\{o,s\}$. Visual and physical histories are advanced by the same action-conditioned predictor $g_\psi$ rather than separate modality-specific predictors. Given a history $Z_t^m$ and an action chunk $a_t$, the predictor produces a future latent that is supervised by the future representations from both modalities. The complete objective is
\begin{equation}
\mathcal{L}_{\text{\ours{}}} = 
\sum_{m,n\in{o,s}}
\left|
g_\psi(Z_t^m,a_t)-z_{t+1}^n
\right|_2^2
+
\beta
\left[
\Omega(z^o)+\Omega(z^s)
\right],
\label{eq:total}
\end{equation}
where actions modulate the predictor through AdaLN~\citep{peebles2023scalable}. The isotropy regularizer $\Omega$ is applied independently to each branch.

The prediction loss contains four source--target terms. The visual-to-visual term is the original LeWM self-prediction objective. The physical-to-physical term learns predictive dynamics within the physical branch. The visual-to-physical term matches a transition predicted from visual history to the corresponding physical future, while the physical-to-visual term provides the reverse constraint. Using both cross-modal directions keeps the coupling symmetric.

Importantly, the two targets associated with a source history describe the same future scene reached under the same action. Thus, $g_\psi(Z_t^o,a_t)$ must match both $z_{t+1}^o$ and $z_{t+1}^s$, and the same requirement applies to predictions from $Z_t^s$. The objective therefore couples the modalities through their action-conditioned evolution, rather than only aligning representations at individual time steps. Sharing $g_\psi$ further encourages histories from both modalities to expose a common transition rule that can be advanced under the same actions.

All target representations are produced by the online encoders, and $f_\theta$, $h_\phi$, and $g_\psi$ are optimized jointly. The physical branch is therefore not a fixed teacher or a predefined physical embedding. It adapts with the visual branch while remaining constrained by the structure of the privileged trajectory. Independent isotropy regularization prevents either branch from satisfying the prediction losses through representational collapse.

\paragraph{A unified privileged-state interface across tasks.}

\begin{figure}[t]
\centering
\includegraphics[width=\linewidth]{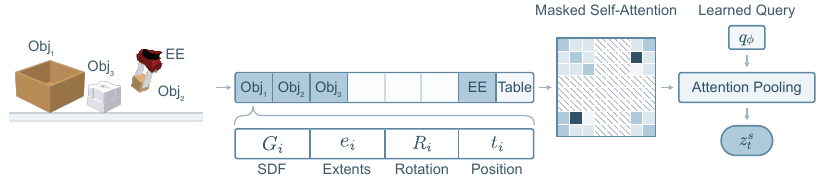}
\caption{\textbf{Unified privileged-state interface.}
A scene is represented by a fixed set of
object slots, together with end-effector (EE) and table tokens; unused
object slots are masked.
Each object token encodes SDF geometry $G_i$, extents $e_i$, rotation $R_i$,
and position $t_i$.
Masked self-attention followed by learned-query attention pooling maps the
resulting token set to the physical latent $z_t^s$.}
\label{fig:privencoding}
\end{figure}

Applying cross-modal grounding across heterogeneous tasks requires a consistent representation of privileged physical state. We therefore describe every scene using a unified rigid-body schema, as shown in Figure~\ref{fig:privencoding}.

Each object contributes a token
\[
\tau_i
=
[\,G_i \,\|\, e_i \,\|\, R_i \,\|\, t_i\,],
\]
where $G_i$ is a fixed-width descriptor of its canonical geometry (a pooled signed-distance field; \citealp{park2019deepsdf}), $e_i$ contains its half-extents, and $(R_i,t_i)$ specifies its pose. An effector token contains the end-effector rotation, position, and finger opening, while a learned table token completes the set. Scenes are padded to six object slots, with unused slots masked from both self-attention and pooling.

Because objects are represented by their geometry rather than their identity, the schema requires neither category labels nor task-specific state fields. A single physical encoder $h_\phi$ can therefore process every configuration in the suite. Appendix~\ref{app:interface} provides the complete dimensions and normalization procedure.

The unified rigid-body schema provides a consistent physical description across configurations. This consistency enables cross-predictive coupling between visual and physical trajectories across tasks, allowing the shared predictor to learn common rigid-body and interaction dynamics without adapting to task-specific state formats.

\label{sec:interface}

\paragraph{Corresponding trajectories as a constraint on predictive evolution.}

The unified interface provides physical histories and futures paired with their visual counterparts at every time step. Corresponding cross-prediction couples this trajectory pairing to predictive evolution: histories from either modality predict future representations in both modalities. In this way, \ours{} constrains how the shared representation geometry evolves under action rather than merely aligning visual and physical states at individual time steps.

\paragraph{Objective decomposition.}
\label{sec:decomp}

The interaction between alignment and predictive evolution becomes explicit under uniform weighting of the four prediction terms. Define
\[
p^m = g_\psi(Z_t^m,a_t),
\qquad
\mu_{t+1}
=
\tfrac{1}{2}
\left(
z_{t+1}^o+z_{t+1}^s
\right).
\]
By the parallelogram identity,
\begin{equation}
\sum_{m,n\in\{o,s\}}
\left\|p^m-z_{t+1}^n\right\|_2^2
=
2\left\|p^o-\mu_{t+1}\right\|_2^2
+
2\left\|p^s-\mu_{t+1}\right\|_2^2
+
\left\|z_{t+1}^o-z_{t+1}^s\right\|_2^2.
\label{eq:decomp}
\end{equation}

Equation~\ref{eq:decomp} exposes two consequences of cross-prediction over correctly paired trajectories. The final term encourages the visual and physical representations of the same future scene to align, while the first two require histories from either modality to predict that shared future under action. \ours{} therefore combines cross-modal representation alignment with an action-conditioned constraint on how the shared latent geometry evolves. Because both targets are online encoder outputs, this coupling is symmetric rather than a one-way distillation toward a fixed physical target.

\paragraph{Deployment.}
\label{sec:planning}

Privileged state is used only during training. At deployment, the physical encoder $h_\phi$ is discarded, leaving the visual encoder and predictor $(f_\theta,g_\psi)$. The deployed model therefore receives the same observations and has the same inference architecture and computational cost as the visual-only baseline.

Candidate action sequences are rolled forward through the learned latent dynamics and scored by the distance between their predicted terminal latent and the encoded goal:
\begin{equation}
J(a_{1:T})
=
\left\|
\hat z_T(a_{1:T}) - f_\theta(o_g)
\right\|_2^2,
\qquad
\hat z_{i+1}
=
g_\psi(\hat Z_i,a_i).
\label{eq:cost}
\end{equation}
The selected sequence is executed, and planning repeats from the resulting state. At test time, the effect of privileged state is carried entirely by the visual representation and predictive dynamics learned during training; privileged state is neither observed nor reconstructed during planning. More forecastable dynamics reduce one source of error in this rollout-based score, although successful control also requires latent distance to remain aligned with physical outcomes.

On the multi-task suite, candidates are drawn using $z$-CEM: a small unconditional variational autoencoder is trained on action chunks from the same corpus, CEM searches its latent space, and each candidate is decoded into a temporally coherent action chunk. Searching this space avoids direct search over raw action sequences, which yields incoherent candidates far from the action distribution seen during training. The autoencoder observes neither the task nor the scene, so the same action prior serves every method. The single-task experiments retain LeWM's released action-space CEM planner. The autoencoder specification and full planner hyperparameters are provided in Appendix~\ref{app:impl}.
\section{Experiments}
\label{sec:results}
\label{sec:experiments}

\paragraph{Setup.}
\label{sec:setup}

We evaluate on four single-task LeWM environments---\emph{Push-T} \citep{chi2023diffusion}, \emph{OGBench-Block} \citep{park2025ogbench}, \emph{Two-Room}, and \emph{Reacher} \citep{tassa2018deepmind}---using the released datasets, planners, and success predicates \citep{maes2026lewm}.
We additionally evaluate on a multi-task tabletop suite built on Meta-World \citep{yu2020meta,todorov2012mujoco}, where a single model is trained jointly across 22 object–task configurations spanning 13 distinct assets. Control is evaluated on six evaluation subfamilies. Forecastability and decodability are measured on held-out episodes spanning every configuration. Appendix~\ref{app:success} gives the exact configuration counts and the mapping between configurations and evaluated families. Core comparisons share the training data, visual encoder, optimization schedule, and deployment planner; each ablation changes only the stated objective or architectural component.

We evaluate three properties.
\emph{Control Utility} is closed-loop task success.
\emph{Forecastability} measures how readily the learned representation supports action-conditioned prediction independently of its co-trained predictor.
For each method, we freeze the visual encoder, discard the co-trained predictor, and fit the same lightweight action-conditioned predictor from scratch. Refitting the predictor separates representation forecastability from encoder–predictor co-adaptation during training. Refitting the predictor directly tests whether predictable transition structure is carried by the learned representation rather than only by the jointly optimized encoder–predictor pair. Because the same predictor family and fitting protocol are used for every frozen representation, differences in rollout error more directly reflect how readily each latent space supports action-conditioned prediction. We report rollout drift relative to a temporal-persistence baseline, with lower values indicating more forecastable dynamics. 
\emph{Decodability} is measured by the $R^2$ of physical variables predicted from the frozen visual latent; we report the position of the manipulated object, and Appendix~\ref{app:forecast} specifies the full decoding target and the remaining groups.
Appendices~\ref{app:success} and~\ref{app:forecast} provide the complete evaluation protocols.

\paragraph{\ours{} improves forecastability.}
\begin{wrapfigure}{r}{0.48\linewidth}
\vspace{-0.3cm}
\centering
\includegraphics[width=\linewidth]{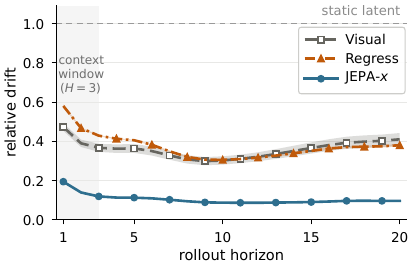}
\caption{\textbf{\ours{} produces more forecastable latent dynamics throughout the rollout.}
Multi-task suite, three training seeds.}
\label{fig:drift}
\vspace{-0.5cm}
\end{wrapfigure}
\label{sec:forecast}
A fresh predictor models \ours{}'s action-conditioned dynamics more accurately than the baselines throughout the rollout. On the multi-task suite, relative
rollout drift falls from $0.361$ for \visual{} to $0.104$ for \ours{}. Figure~\ref{fig:drift} shows that the separation persists at every horizon, including beyond $h=3$, when the rollout becomes fully autoregressive. Similar temporal-persistence errors and rank-matched PCA show that neither reduced temporal variation nor effective rank alone explains the gap (Appendix~\ref{app:forecast}).

The improvement is consistent across all four matched single-task environments (Table~\ref{tab:lewm-probe}). Relative drift decreases from $0.503$ to $0.221$ on Two-Room, from $0.507$ to $0.269$ on OGBench-Block, from $0.399$ to $0.258$ on Push-T, and from $0.313$ to $0.233$ on Reacher.

\begin{table}[t]
\caption{\textbf{\ours{} produces the most forecastable visual dynamics across all four matched single-task environments.} Forecastability is measured using a predictor fitted from scratch to each frozen representation. Object-position $R^2$ measures how well the environment's object position is decoded from the frozen visual latent. Values are means $\pm$ sample SD over three training seeds. Bold marks the best mean in each comparison; exact ties are all bold.}
\label{tab:lewm-probe}
\centering
\small
\setlength{\tabcolsep}{2.6pt}
\begin{tabular}{lccc@{\hskip 7pt}ccc}
\toprule
& \multicolumn{3}{c}{Relative drift $\downarrow$}
& \multicolumn{3}{c}{Object-position $R^2$ $\uparrow$} \\
Environment & \ours{} & \visual{} & \regress{}
& \ours{} & \visual{} & \regress{} \\
\midrule
Two-Room
& \best{$0.221_{\pm0.021}$} & $0.503_{\pm0.010}$ & $0.480_{\pm0.008}$
& \best{$0.998_{\pm0.000}$} & $0.921_{\pm0.011}$ & $0.936_{\pm0.007}$ \\
OGBench-Block
& \best{$0.269_{\pm0.007}$} & $0.507_{\pm0.026}$ & $0.393_{\pm0.018}$
& $0.982_{\pm0.003}$ & $0.879_{\pm0.009}$ & \best{$0.983_{\pm0.001}$} \\
Push-T
& \best{$0.258_{\pm0.007}$} & $0.399_{\pm0.118}$ & $0.343_{\pm0.039}$
& \best{$0.991_{\pm0.001}$} & $0.944_{\pm0.009}$ & $0.968_{\pm0.007}$ \\
Reacher
& \best{$0.233_{\pm0.009}$} & $0.313_{\pm0.009}$ & $0.309_{\pm0.003}$
& \best{$0.999_{\pm0.000}$} & \best{$0.999_{\pm0.000}$} & \best{$0.999_{\pm0.000}$} \\
\bottomrule
\end{tabular}
\end{table}

\paragraph{\ours{} improves multi-task control.}
\label{sec:main}

\begin{figure}[tb]
\centering
\includegraphics[width=\linewidth]{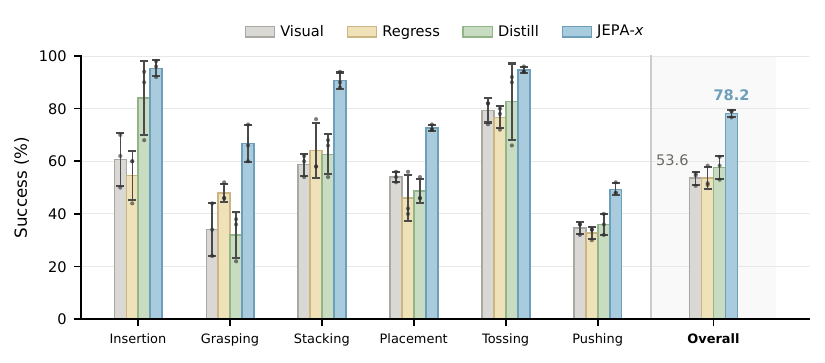}
\caption{\textbf{Grounding improves multi-task control.}
\ours{} raises overall success from $53.6\%$ to $78.2\%$, with improvements across all six interaction families.
Regressing a comprehensive pose-based physical-state target from the visual latent does not recover the gain.
Bars show means $\pm$ sample SD over three training seeds; dots show individual seeds.}
\label{fig:multitask-control}
\vspace{-0.25cm}
\end{figure}

On the multi-task suite, \ours{} raises mean task success from $53.6\%$ to $78.2\%$, a gain of $24.6$ percentage points under the same planner, action prior, and candidate budget (Figure~\ref{fig:multitask-control}). The improvement holds across all six interaction families.

\begin{figure}[tb]
\centering
\includegraphics[width=\linewidth]{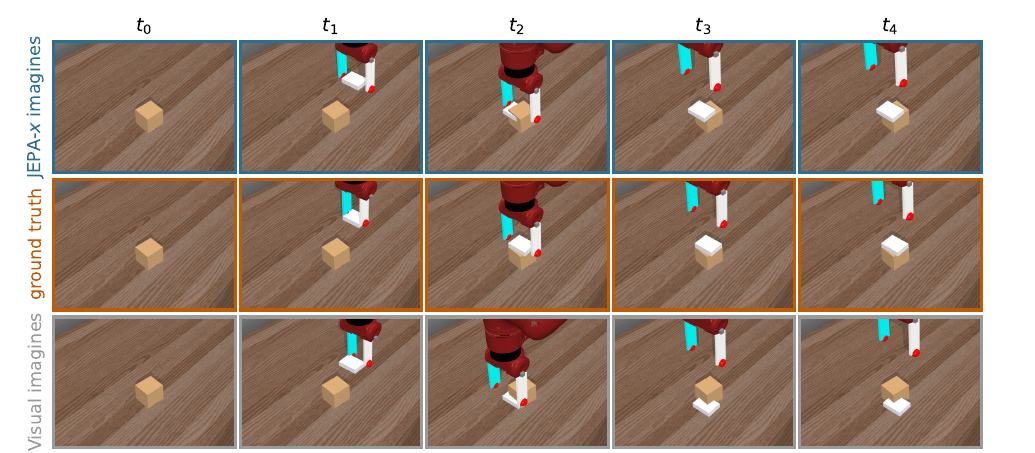}
\caption{\textbf{Rolled over identical expert actions, \ours{} imagines the eraser reaching the block; \visual{} imagines it beside the block.}
Middle row: ground truth. Top and bottom: each model's own rollout over the same action chunks from the same observation, with the decoded object and end-effector positions written back into the simulator and rendered.}
\label{fig:imagined}
\end{figure}

Figure~\ref{fig:imagined} compares the two models' predictions under identical action sequences. Because both models predict in latent space, we fit a separate state decoder, decode the manipulated-object and end-effector positions, and render the resulting rollouts alongside the ground truth. \ours{} imagines the eraser arriving on top of the block, whereas \visual{} places it on the table beside the block.
Across the three planning cycles of this episode, the decoded terminal position error averages $1.4$~cm for \ours{} and $3.1$~cm for \visual{} (Appendix~\ref{app:imagination}).
The decoded rollouts make the drift difference concrete: the same actions lead the two models to different imagined terminal configurations.

State regression does not recover these forecastability or control gains. \regress{} directly predicts a comprehensive pose-based target from the visual latent and raises manipulated-object position decodability to $R^2=0.991$, from $0.978$ for both \ours{} and \visual{}. Its control success remains at $53.7\%$ and its relative rollout drift at $0.373$. Position can therefore be readily available to a probe even when the latent transition remains difficult to forecast. The same holds when the privileged head is asked for the physical \emph{future} rather than the present: an action-conditioned head predicting the next state, or its increment, from $(z^o_t,a_t)$ leaves control at $54.4\%$ and $51.3\%$ and relative drift at $0.386$ and $0.359$, all within the seed spread of \visual{} (Appendix~\ref{app:futaux}). What distinguishes \ours{} is therefore not that the target lies in the future, but that it lies in the physical branch's learned representation.
On the matched single-task environments, \ours{} significantly improves control on Two-Room and OGBench-Block, while its mean success is within one percentage point of \visual{} on Push-T and Reacher (Table~\ref{tab:lewm-control}). Forecastability improves across all four environments, whereas the control gains are task dependent. This separation helps clarify the role of forecastability in planning. \ours{} makes the latent dynamics easier to predict in all four environments, but the downstream control benefit remains task dependent. This pattern is consistent with rollout error mattering more when it changes the planner's ranking of candidate actions, although the present evaluation does not isolate that mechanism. The single-task results therefore support a more limited claim: forecastable dynamics improve one component required by rollout-based planning, rather than guaranteeing higher control success on every task.

\begin{table}[t]
\caption{\textbf{Matched single-task control success.}
Values are mean success rates (\%) $\pm$ sample SD over three training seeds.
Bold marks the best mean in each row; exact ties are all bold.}
\label{tab:lewm-control}
\centering
\small
\setlength{\tabcolsep}{8pt}
\begin{tabular}{lccc}
\toprule
Environment & \ours{} & \visual{} & \regress{} \\
\midrule
Two-Room
& \best{$94.8_{\pm1.0}$} & $74.7_{\pm4.1}$ & $75.8_{\pm3.8}$ \\
OGBench-Block
& \best{$78.7_{\pm2.7}$} & $64.9_{\pm1.4}$ & $70.2_{\pm0.8}$ \\
Push-T
& $94.1_{\pm0.4}$ & $94.9_{\pm1.1}$ & \best{$96.9_{\pm2.7}$} \\
Reacher
& \best{$83.6_{\pm1.4}$} & $82.7_{\pm2.1}$ & $82.6_{\pm0.4}$ \\
\bottomrule
\end{tabular}
\end{table}

\paragraph{Correspondence and cross-prediction play distinct roles.}
\label{sec:abl}
\label{sec:abl-mech}

We next separate the effects of correct visual–physical trajectory pairing, direct cross-modal prediction, and predictor sharing. \textsc{Cross-only} retains cross-prediction but uses separate predictors for the visual and physical streams. \textsc{Share-only} retains the shared predictor but removes the cross-modal prediction terms. \textsc{Align-only} replaces cross-modal prediction with symmetric frame-level alignment while retaining within-modality prediction. \distill{} replaces predictive coupling with pointwise regression onto a stop-gradient physical latent. Finally, \shuffle{} retains the complete \ours{} architecture but uses a fixed corpus-wide derangement, assigning each visual trajectory a single intact but incorrect privileged partner throughout training.

\begin{table}[t]
\caption{\textbf{Correct trajectory pairing supports control, while direct cross-modal prediction improves forecastability beyond alignment.}
The design columns indicate cross-modal prediction, predictor sharing, and correct visual--physical pairing (\yes{} present, \perm{} present but mispaired, \no{} absent).
Drift and control are means $\pm$ sample SD over three training seeds; bold marks the best mean in each column.}
\label{tab:mech}
\centering
\small
\setlength{\tabcolsep}{6pt}
\begin{tabular}{lccc@{\hskip 10pt}cc}
\toprule
Variant & Cross-pred. & Shared & Corresp. & Drift $\downarrow$ & Control $\uparrow$ \\
\midrule
\ours{}
& \yes & \yes & \yes
& $0.104_{\pm0.003}$
& \best{$78.2_{\pm1.3}$} \\
\textsc{Cross-only}
& \yes & \no & \yes
& \best{$0.101_{\pm0.001}$}
& $73.0_{\pm0.9}$ \\
\textsc{Share-only}
& \no & \yes & \no
& $0.309_{\pm0.020}$
& $56.7_{\pm3.5}$ \\
\textsc{Align-only}
& \no & \yes & \yes
& $0.158_{\pm0.001}$
& $72.6_{\pm3.6}$ \\
\distill{}
& \no & \no & \yes
& $0.516_{\pm0.004}$
& $57.7_{\pm4.4}$ \\
\shuffle{}
& \yes & \yes & \perm
& $0.540_{\pm0.023}$
& $3.6_{\pm1.3}$ \\
\visual{}
& \no & \no & \no
& $0.361_{\pm0.024}$
& $53.6_{\pm2.5}$ \\
\bottomrule
\end{tabular}
\end{table}

Table~\ref{tab:mech} shows that cross-modal predictive coupling remains effective without predictor sharing, whereas predictor sharing alone does not recover the control gain. \textsc{Cross-only} reaches $73.0\%$ control, recovering roughly four fifths of the gain from \visual{} to \ours{}, with comparable rollout drift to \ours{}. \textsc{Share-only} reaches $56.7\%$ control, close to the \visual{} baseline.
Most of the control gain therefore survives without predictor-parameter sharing. Cross-modal predictive coupling remains effective with separate predictors, whereas sharing predictor parameters without cross-modal prediction produces only a small improvement over \visual{}.

\textsc{Align-only} recovers most of the control gain ($72.6\%$ against $78.2\%$) but has higher drift than \ours{} ($0.158$ against $0.104$), indicating that direct cross-modal prediction improves forecastability beyond pointwise alignment. Its aligned states are also trained with within-modality prediction, so it still carries cross-modal information forward indirectly.
Across these variants, objectives using correctly paired visual--physical trajectories through alignment or cross-prediction retain most of the control improvement, while direct cross-modal prediction produces substantially lower rollout drift than frame-level alignment.

Pointwise distillation does not reproduce \ours{}'s gains: \distill{} reaches $57.7\%$ control and increases relative rollout drift to $0.516$, compared with $0.361$ for \visual{}. Together with \regress{}, this shows that making physical state decodable or matching it pointwise is not a substitute for grounding the transition itself.

\shuffle{} tests the importance of correct visual–physical correspondence while preserving a stable alternative pairing: each visual trajectory is assigned one intact but incorrect privileged trajectory throughout training. Despite retaining the full predictive architecture and intact privileged trajectories, this fixed mispairing degrades both properties: control falls to $3.6\%$ and relative drift rises to $0.540$, above the $0.361$ of \visual{}, which receives no privileged grounding. A consistent but incorrect pairing is therefore more damaging than omitting privileged grounding in this construction. Training against incompatible cross-modal targets disrupts the transition structure available to a freshly fitted predictor, rather than leaving the visual representation unaffected.

These ablations identify distinct contributions within cross-predictive grounding. Correct trajectory correspondence is necessary for cross-predictive grounding to improve rather than disrupt the visual representation: replacing the corresponding physical trajectory with a fixed incorrect partner degrades both forecastability and control below the ungrounded baseline. Given correct pairs, direct cross-modal prediction shapes action-conditioned evolution more effectively than frame-level alignment, reducing relative drift from \(0.158\) to approximately \(0.10\). Predictor sharing alone recovers neither benefit, while pointwise regression and distillation show that physical-state readout or latent matching is not a substitute for predictive coupling. Together, the results support the claim that the gain comes from constraining visual latent transitions with the corresponding physical future, rather than merely adding physical information or sharing model parameters.

\section{Discussion and Conclusion}
\label{sec:discussion}

Across the multi-task suite, \ours{} improves both fresh-predictor forecastability and closed-loop control, while direct state regression increases position decodability without comparable gains in either. Snapshot physical content therefore does not determine whether a representation supports reliable rollout prediction. \ours{} instead couples visual latent evolution to corresponding physical trajectories. Because both encoders are learned jointly, this does not impose a canonical physical representation; privileged state provides a structured view of the same action-conditioned trajectory that constrains predictive evolution.

The ablations distinguish correct correspondence from the form of cross-modal coupling. Replacing each corresponding physical trajectory with a fixed incorrect partner degrades forecastability and control below the visual-only baseline, showing that correct pairing is essential for cross-predictive grounding. Given correct pairs, direct cross-modal prediction reduces rollout drift beyond frame-level alignment, whereas predictor sharing alone recovers little of the gain. Pointwise regression and distillation likewise fail to reproduce the forecastability improvement, locating the benefit in corresponding predictive coupling rather than parameter sharing or snapshot-level matching.

This distinction matters for rollout-based planning, where candidate actions are ranked by distances between predicted terminal latents and the goal. \ours{} retains its lower drift after the co-trained predictor is replaced, indicating that the learned representation carries more forecastable transition structure. However, the single-task results show that improved forecastability does not guarantee higher success in every environment. It strengthens one component required for rollout-based planning, while its control benefit remains task dependent.

Our results are limited to simulation, paired privileged trajectories during training, and a fixed rigid-body state schema. Extending the approach to real-world observations and broader physical interactions remains future work.

\label{sec:conclusion}

\bibliography{references}

@inproceedings{ha2018world,
  title={Recurrent world models facilitate policy evolution},
  author={Ha, David and Schmidhuber, J{\"u}rgen},
  booktitle={Advances in Neural Information Processing Systems},
  volume={31},
  year={2018}
}

@inproceedings{hafner2020dream,
  title={Dream to control: Learning behaviors by latent imagination},
  author={Hafner, Danijar and Lillicrap, Timothy and Ba, Jimmy and Norouzi, Mohammad},
  booktitle={International Conference on Learning Representations},
  year={2020}
}

@article{hafner2023mastering,
  title={Mastering diverse control tasks through world models},
  author={Hafner, Danijar and Pasukonis, Jurgis and Ba, Jimmy and Lillicrap, Timothy},
  journal={Nature},
  volume={640},
  pages={647--653},
  year={2025},
  doi={10.1038/s41586-025-08744-2}
}

@inproceedings{hansen2024tdmpc2,
  title={{TD-MPC2}: Scalable, robust world models for continuous control},
  author={Hansen, Nicklas and Su, Hao and Wang, Xiaolong},
  booktitle={International Conference on Learning Representations},
  year={2024}
}

@article{zhou2024dinowm,
  title={{DINO-WM}: World models on pre-trained visual features enable zero-shot planning},
  author={Zhou, Gaoyue and Pan, Hengkai and LeCun, Yann and Pinto, Lerrel},
  journal={arXiv preprint arXiv:2411.04983},
  year={2024}
}

@article{maes2026lewm,
  title={{LeWorldModel}: Stable end-to-end joint-embedding predictive architecture from pixels},
  author={Maes, Lucas and Lidec, Quentin Le and Scieur, Damien and LeCun, Yann and Balestriero, Randall},
  journal={arXiv preprint arXiv:2603.19312},
  year={2026}
}

@article{sobal2025pldm,
  title={Learning from reward-free offline data: A case for planning with latent dynamics models},
  author={Sobal, Uladzislau and Zhang, Wancong and Cho, Kyunghyun and Balestriero, Randall and Rudner, Tim G. J. and LeCun, Yann},
  journal={Advances in Neural Information Processing Systems},
  volume={38},
  pages={43905--43941},
  year={2025}
}

@inproceedings{chen2020learning,
  title={Learning by cheating},
  author={Chen, Dian and Zhou, Brady and Koltun, Vladlen and Kr{\"a}henb{\"u}hl, Philipp},
  booktitle={Proceedings of the Conference on Robot Learning},
  pages={66--75},
  volume={100},
  year={2020},
  organization={PMLR}
}

@inproceedings{kumar2021rma,
  title={{RMA}: Rapid motor adaptation for legged robots},
  author={Kumar, Ashish and Fu, Zipeng and Pathak, Deepak and Malik, Jitendra},
  booktitle={Proceedings of Robotics: Science and Systems},
  year={2021},
  doi={10.15607/RSS.2021.XVII.011}
}

@inproceedings{pinto2018asymmetric,
  title={Asymmetric actor critic for image-based robot learning},
  author={Pinto, Lerrel and Andrychowicz, Marcin and Welinder, Peter and Zaremba, Wojciech and Abbeel, Pieter},
  booktitle={Proceedings of Robotics: Science and Systems},
  year={2018},
  doi={10.15607/RSS.2018.XIV.008}
}

@misc{lecun2022path,
  title={A path towards autonomous machine intelligence},
  author={LeCun, Yann},
  howpublished={OpenReview},
  note={Version 0.9.2},
  year={2022}
}

@inproceedings{assran2023self,
  title={Self-supervised learning from images with a joint-embedding predictive architecture},
  author={Assran, Mahmoud and Duval, Quentin and Misra, Ishan and Bojanowski, Piotr and Vincent, Pascal and Rabbat, Michael and LeCun, Yann and Ballas, Nicolas},
  booktitle={2023 IEEE/CVF Conference on Computer Vision and Pattern Recognition (CVPR)},
  pages={15619--15629},
  year={2023},
  organization={IEEE}
}

@article{bardes2023mcjepa,
  title={{MC-JEPA}: A joint-embedding predictive architecture for self-supervised learning of motion and content features},
  author={Bardes, Adrien and Ponce, Jean and LeCun, Yann},
  journal={arXiv preprint arXiv:2307.12698},
  year={2023}
}

@article{bardes2024vjepa,
  title={Revisiting feature prediction for learning visual representations from video},
  author={Bardes, Adrien and Garrido, Quentin and Ponce, Jean and Chen, Xinlei and Rabbat, Michael and LeCun, Yann and Assran, Mahmoud and Ballas, Nicolas},
  journal={arXiv preprint arXiv:2404.08471},
  year={2024}
}

@article{balestriero2025lejepa,
  title={{LeJEPA}: Provable and scalable self-supervised learning without the heuristics},
  author={Balestriero, Randall and LeCun, Yann},
  journal={arXiv preprint arXiv:2511.08544},
  year={2025}
}

@inproceedings{dosovitskiy2021image,
  title={An image is worth 16x16 words: Transformers for image recognition at scale},
  author={Dosovitskiy, Alexey and Beyer, Lucas and Kolesnikov, Alexander and Weissenborn, Dirk and Zhai, Xiaohua and Unterthiner, Thomas and Dehghani, Mostafa and Minderer, Matthias and Heigold, Georg and Gelly, Sylvain and others},
  booktitle={International Conference on Learning Representations},
  year={2021}
}

@inproceedings{peebles2023scalable,
  title={Scalable diffusion models with transformers},
  author={Peebles, William and Xie, Saining},
  booktitle={2023 IEEE/CVF International Conference on Computer Vision (ICCV)},
  pages={4172--4182},
  year={2023},
  organization={IEEE}
}

@inproceedings{yu2020meta,
  title={{Meta-World}: A benchmark and evaluation for multi-task and meta reinforcement learning},
  author={Yu, Tianhe and Quillen, Deirdre and He, Zhanpeng and Julian, Ryan and Hausman, Karol and Finn, Chelsea and Levine, Sergey},
  booktitle={Proceedings of the Conference on Robot Learning},
  pages={1094--1100},
  volume={100},
  year={2020},
  organization={PMLR}
}

@inproceedings{gupta2016cross,
  title={Cross modal distillation for supervision transfer},
  author={Gupta, Saurabh and Hoffman, Judy and Malik, Jitendra},
  booktitle={Proceedings of the IEEE Conference on Computer Vision and Pattern Recognition},
  pages={2827--2836},
  year={2016}
}

@inproceedings{tian2020contrastive,
  title={Contrastive multiview coding},
  author={Tian, Yonglong and Krishnan, Dilip and Isola, Phillip},
  booktitle={European Conference on Computer Vision},
  pages={776--794},
  year={2020},
  organization={Springer}
}

@inproceedings{park2025ogbench,
  title={{OGBench}: Benchmarking offline goal-conditioned {RL}},
  author={Park, Seohong and Frans, Kevin and Eysenbach, Benjamin and Levine, Sergey},
  booktitle={International Conference on Learning Representations},
  year={2025}
}

@article{tassa2018deepmind,
  title={{DeepMind} Control Suite},
  author={Tassa, Yuval and Doron, Yotam and Muldal, Alistair and Erez, Tom and Li, Yazhe and Casas, Diego de Las and Budden, David and Abdolmaleki, Abbas and Merel, Josh and Lefrancq, Andrew and others},
  journal={arXiv preprint arXiv:1801.00690},
  year={2018}
}

@article{chi2023diffusion,
  title={Diffusion policy: Visuomotor policy learning via action diffusion},
  author={Chi, Cheng and Xu, Zhenjia and Feng, Siyuan and Cousineau, Eric and Du, Yilun and Burchfiel, Benjamin and Tedrake, Russ and Song, Shuran},
  journal={The International Journal of Robotics Research},
  volume={44},
  number={10-11},
  pages={1684--1704},
  year={2025},
  publisher={Sage Publications Sage UK: London, England}
}

@inproceedings{todorov2012mujoco,
  title={{MuJoCo}: A physics engine for model-based control},
  author={Todorov, Emanuel and Erez, Tom and Tassa, Yuval},
  booktitle={2012 IEEE/RSJ International Conference on Intelligent Robots and Systems},
  pages={5026--5033},
  year={2012},
  organization={IEEE}
}

@inproceedings{park2019deepsdf,
  title={{DeepSDF}: Learning continuous signed distance functions for shape representation},
  author={Park, Jeong Joon and Florence, Peter and Straub, Julian and Newcombe, Richard and Lovegrove, Steven},
  booktitle={2019 IEEE/CVF Conference on Computer Vision and Pattern Recognition (CVPR)},
  pages={165--174},
  year={2019},
  organization={IEEE}
}

@article{vapnik2015learning,
  title={Learning using privileged information: similarity control and knowledge transfer},
  author={Vapnik, Vladimir and Izmailov, Rauf},
  journal={The Journal of Machine Learning Research},
  volume={16},
  number={1},
  pages={2023--2049},
  year={2015},
  publisher={JMLR. org}
}

@article{lee2020learning,
  title={Learning quadrupedal locomotion over challenging terrain},
  author={Lee, Joonho and Hwangbo, Jemin and Wellhausen, Lorenz and Koltun, Vladlen and Hutter, Marco},
  journal={Science Robotics},
  volume={5},
  number={47},
  pages={eabc5986},
  year={2020},
  publisher={American Association for the Advancement of Science}
}

@inproceedings{hafner2019learning,
  title={Learning latent dynamics for planning from pixels},
  author={Hafner, Danijar and Lillicrap, Timothy and Fischer, Ian and Villegas, Ruben and Ha, David and Lee, Honglak and Davidson, James},
  booktitle={International Conference on Machine Learning},
  pages={2555--2565},
  year={2019},
  organization={PMLR}
}

@article{schrittwieser2020mastering,
  title={Mastering {Atari}, {Go}, chess and shogi by planning with a learned model},
  author={Schrittwieser, Julian and Antonoglou, Ioannis and Hubert, Thomas and Simonyan, Karen and Sifre, Laurent and Schmitt, Simon and Guez, Arthur and Lockhart, Edward and Hassabis, Demis and Graepel, Thore and others},
  journal={Nature},
  volume={588},
  number={7839},
  pages={604--609},
  year={2020},
  publisher={Nature Publishing Group UK London}
}

@inproceedings{yamada2024twist,
  title={{TWIST}: Teacher-student world model distillation for efficient sim-to-real transfer},
  author={Yamada, Jun and Rigter, Marc and Collins, Jack and Posner, Ingmar},
  booktitle={2024 IEEE International Conference on Robotics and Automation (ICRA)},
  pages={9190--9196},
  year={2024},
  organization={IEEE}
}

@inproceedings{hu2024privileged,
  title={Privileged sensing scaffolds reinforcement learning},
  author={Hu, Edward and Springer, James and Rybkin, Oleh and Jayaraman, Dinesh},
  booktitle={International Conference on Learning Representations},
  year={2024}
}

@article{huang2025pigdreamer,
  title={{PIGDreamer}: Privileged information guided world models for safe partially observable reinforcement learning},
  author={Huang, Dongchi and Wang, Jiaqi and Li, Yang and Xia, Chunhe and Zhang, Tianle and Zhang, Kaige},
  journal={arXiv preprint arXiv:2508.02159},
  year={2025}
}

@article{kim2026pri4r,
  title={{Pri4R}: Learning world dynamics for vision-language-action models with privileged {4D} representation},
  author={Kim, Jisoo and Cho, Jungbin and Chu, Sanghyeok and Bal, Ananya and Kim, Jinhyung and Lee, Gunhee and Lee, Sihaeng and Kim, Seung Hwan and Han, Bohyung and Lee, Hyunmin and others},
  journal={arXiv preprint arXiv:2603.01549},
  year={2026}
}

@article{nie2026physjepa,
  title={{Phys-JEPA}: Physics-informed latent world models for multivariate time-series forecasting},
  author={Nie, Weizhi and Liu, Weichao and Guo, Honglin and Su, Yuting},
  journal={arXiv preprint arXiv:2606.16076},
  year={2026}
}

@article{zeng2026phylatent,
  title={{PhyLatent}: Learning dynamics-relevant representations for {JEPA} world models},
  author={Zeng, Xi and Ren, Haojie and Song, Ziying},
  journal={arXiv preprint arXiv:2608.05720},
  year={2026}
}
\bibliographystyle{iclr2027_conference}

\appendix
\clearpage
\appendix

\section{Suite and Evaluation Details}
\label{app:success}

\paragraph{Suite composition.}
Our Meta-World-based corpus contains $22$ object--task configurations over $13$ distinct assets, with $450$ clean episodes per configuration before noise augmentation. A single model is trained jointly across all configurations.

\begin{table}[ht]
\caption{The $22$ configurations in the multi-task suite.}
\label{tab:suite-configs}
\centering
\small
\setlength{\tabcolsep}{4pt}
\begin{tabular}{lp{0.72\linewidth}}
\toprule
Interaction group & Configurations \\
\midrule
Grasp--move--place &
battery, block, can, tennis ball \\

Peg / plug insertion &
peg-align-insert, peg-insert-pull, plug-insert \\

Planar pushing &
block, book, eraser \\

Stacking &
battery-on-phone, block-on-book, eraser-on-block \\

Righting / toppling &
stand-up-can, stand-up-thermos, topple-battery,
topple-can, topple-thermos \\

Tossing &
battery, block, eraser, tennis ball \\
\bottomrule
\end{tabular}
\end{table}

\paragraph{Success predicates.}
Success is determined from the executed physical trajectory rather than latent distance to the goal. Positional tolerances are asset-specific, and orientation is evaluated modulo each asset's declared symmetry.

\begin{table}[ht]
\caption{Success predicates for the six evaluated interaction families.}
\label{tab:success-predicates}
\centering
\small
\setlength{\tabcolsep}{5pt}
\begin{tabular}{lp{0.72\linewidth}}
\toprule
Family & Success condition \\
\midrule
Grasping &
Object at least $5$\,cm above its initial table height over the final
$1$\,s, with in-grasp slip at most $1$\,cm. \\

Stacking &
Support contact between the stacked object and its base after a
$2$\,s settle window, with the base orientation preserved. \\

Insertion &
Insertion depth at least $90\%$ of the target cavity depth,
axis tilt at most $3^\circ$, and gripper open at termination. \\

Placement &
Object at rest within the asset-specific positional tolerance of the target,
with orientation compared modulo symmetry. \\

Pushing &
Object within the asset-specific tolerance of the planar goal,
with orientation compared to the initial pose modulo symmetry. \\

Tossing &
Object inside the container interior at the end of the trajectory. \\
\bottomrule
\end{tabular}
\end{table}

\paragraph{Scenario sampling.}
Each evaluated interaction family contains $50$ held-out scenarios per training seed, giving $300$ scenarios per model. Scenarios are stratified across configuration--subtask pairs and sampled with a fixed seed, so all methods are evaluated from the same initial states. The $300$ scenarios are drawn from $266$ source episodes; statistical tests therefore cluster segments originating from the same episode.

The interaction groups in Table~\ref{tab:suite-configs} label corpus configurations, whereas the six evaluated families are subtask labels annotated within each episode. Scenarios are keyed by configuration--subtask pair, and each family's quota is filled round-robin across every configuration containing that subtask. A grasp--move--place configuration therefore contributes segments to the grasping and placement families, the peg configurations to insertion, and the pushing, stacking and tossing configurations to their like-named families.

For the four LeWM environments, we use the benchmark's released success predicates and evaluation loop unchanged. Each controller is evaluated on the same $N=150$ paired scenarios.

\section{Full Control Results}
\label{app:evaldetails}

\paragraph{Per-family results.}
Table~\ref{tab:family-control} reports the per-family results underlying Figure~\ref{fig:multitask-control}.

\begin{table}[ht]
\caption{Per-family control success (\%), mean $\pm$ sample SD over three training seeds. Families are ordered by the \ours{}--\visual{} gap, matching Figure~\ref{fig:multitask-control}.}
\label{tab:family-control}
\centering
\small
\setlength{\tabcolsep}{4.5pt}
\begin{tabular}{lccccc}
\toprule
Family & \ours{} & \visual{} & \regress{} & \distill{} & \shuffle{} \\
\midrule
Insertion & $95.3_{\pm3.1}$ & $60.7_{\pm10.1}$ & $54.7_{\pm9.2}$ & $84.0_{\pm14.0}$ & $16.0_{\pm6.0}$ \\
Grasping  & $66.7_{\pm7.0}$ & $34.0_{\pm10.0}$ & $48.0_{\pm3.5}$ & $32.0_{\pm8.7}$ & $1.3_{\pm1.2}$ \\
Stacking  & $90.7_{\pm3.1}$ & $58.7_{\pm4.2}$ & $64.0_{\pm10.4}$ & $62.7_{\pm7.6}$ & $0.7_{\pm1.2}$ \\
Placement & $72.7_{\pm1.2}$ & $54.0_{\pm2.0}$ & $46.0_{\pm8.7}$ & $48.7_{\pm4.6}$ & $2.0_{\pm2.0}$ \\
Tossing   & $94.7_{\pm1.2}$ & $79.3_{\pm4.6}$ & $76.7_{\pm4.2}$ & $82.7_{\pm14.5}$ & $1.3_{\pm2.3}$ \\
Pushing   & $49.3_{\pm2.3}$ & $34.7_{\pm2.3}$ & $32.7_{\pm2.3}$ & $36.0_{\pm4.0}$ & $0.0_{\pm0.0}$ \\
\midrule
All six   & \best{$78.2_{\pm1.3}$}
          & $53.6_{\pm2.5}$
          & $53.7_{\pm4.1}$
          & $57.7_{\pm4.4}$
          & $3.6_{\pm1.3}$ \\
\bottomrule
\end{tabular}
\end{table}

\paragraph{Statistical tests.}
For each training seed, suite comparisons use an episode-clustered paired sign-flip test with $10^5$ Monte Carlo assignments.
Scenarios originating from the same source episode share a sign.
The per-seed \ours{}--\visual{} differences are $+28.3$, $+22.0$, and $+23.7$ percentage points, while the respective \ours{}--\distill{} differences are $+17.3$, $+18.3$, and $+26.0$ percentage points.
All six seed-level comparisons yield $p_{\mathrm{MC}}\leq 10^{-5}$, the smallest value $10^5$ assignments can resolve.
These tests quantify paired scenario-level differences for each trained model; consistency across training seeds is reported separately.

Single-task comparisons use exact McNemar tests on the same $N=150$ paired scenarios.
\ours{} exceeds \visual{} in every seed on both Two-Room and OGBench-Block, with the largest $p$-value equal to $0.023$.

For the comparison between \ours{} and \textsc{Cross-only}, the training run is the unit of analysis.
Mean suite control is $72.3/74.0/72.7$ for \textsc{Cross-only} and $79.0/76.7/79.0$ for \ours{}.

\section{Forecastability Protocol and Additional Results}
\label{app:forecast}

\paragraph{Fresh-predictor probe.}
To measure representation forecastability independently of the transition model used during representation learning, we freeze each encoder, cache its latents, discard the original predictor, and train a new predictor from scratch. We use the same lightweight probe family across methods, equalizing predictor capacity so that differences primarily reflect predictive structure in the learned representation rather than the co-trained transition model.

For the single-task experiments, the probe is action-conditioned. Its input concatenates the $H=3$ most recent latent frames with the aligned $H=3$ action chunks. A three-layer MLP of width $512$ with GELU activations predicts the next latent as a residual update to the most recent frame. Latents are standardized per dimension using statistics from the fitting episodes; actions are used as stored. We train with AdamW at learning rate $10^{-3}$ and weight decay $10^{-5}$ for $30{,}000$ steps, with batch size $256$ on $1000$ episodes and a randomized fit/evaluation episode split.

At evaluation, the probe is rolled out autoregressively for $20$ steps, feeding back its own predictions while consuming the recorded action sequence. The resulting error therefore measures prediction under actions rather than latent smoothness alone. Normalization by a constant-latent baseline further removes credit for temporal persistence.

The multi-task results in Table~\ref{tab:suite-forecast} and Figure~\ref{fig:drift} use the same probe family with a direct rather than residual output head, $20{,}000$ training steps, batch size $512$, cosine learning-rate decay, and an episode-level fit/evaluation split.

Probe fitting and evaluation always use disjoint episodes. Relative drift is computed independently at each rollout horizon and normalized by a constant-latent predictor:
\begin{equation}
\mathrm{drift}(h)=
\frac{
\mathbb{E}\lVert \hat z_{t+h}-z_{t+h}\rVert_2
}{
\mathbb{E}\lVert z_t-z_{t+h}\rVert_2
}.
\end{equation}
The reported aggregate is the mean of these per-horizon ratios over
$h=1,\ldots,20$.

\paragraph{Manipulated-object decoding probe.}
Decodability is measured using the same frozen encoder checkpoints and episode-level split as the dynamics probe, so both columns of Table~\ref{tab:suite-forecast} evaluate the same representations under matched data partitions.
A single three-layer MLP of width $512$ with GELU activations maps one frozen latent frame to physical variables; it is trained with AdamW at learning rate $10^{-3}$ and weight decay $10^{-5}$ for $6000$ steps at batch size $1024$, and evaluated on the held-out episodes.
The decoding target is a $15$-dimensional vector comprising the manipulated object's normalized position, its orientation as a flattened $3\times3$ rotation matrix, and the normalized end-effector position.
We compute $R^2 = 1 - \mathrm{SSE}/\mathrm{SST}$ separately for each target group over all held-out frames, with the total sum of squares taken about the held-out mean.
The tables report $R^2$ for the object-position group.
Canonical geometry, bounding-box extents, and masked object slots are not included in the decoding target.

\paragraph{Multi-task forecastability.}
Table~\ref{tab:suite-forecast} reports the quantities underlying the multi-task comparison.
Under the controlled fresh-predictor protocol, JEPA-x has lower raw rollout error and normalized drift than the visual-only and state-regression baselines. \regress{} and \distill{} make manipulated-object position more decodable, but neither substantially reduces rollout drift relative to \visual{}. In contrast, \ours{} reduces drift without increasing the reported position decodability over \visual{}. \shuffle{} increases drift beyond every other variant and collapses control, showing that incorrect visual--physical pairing degrades forecastability and control together. Decodability and forecastability are therefore distinct properties, and the ordering of one does not predict the ordering of the other.

\begin{table}[ht]
\caption{Forecastability and manipulated-object position decodability on the multi-task suite. Both metrics use the same frozen encoders and held-out split, with a separately fitted probe for each metric. Drift and $R^2$ are means $\pm$ sample SD over three training seeds; bold marks the best reported mean. Raw and Copy are averaged over horizons and seeds; Drift is the mean of the per-horizon ratios and therefore differs from their ratio.}
\label{tab:suite-forecast}
\centering
\small
\setlength{\tabcolsep}{5pt}
\begin{tabular}{lcccc}
\toprule
Arm &
Raw $\downarrow$ &
Copy &
Drift $\downarrow$ &
Manip.-object pos.\ $R^2$ $\uparrow$ \\
\midrule
\ours{} &
1.18 & 12.17 &
\best{$0.104_{\pm0.003}$} &
$0.978_{\pm0.002}$ \\

\visual{} &
4.43 & 12.42 &
$0.361_{\pm0.024}$ &
$0.978_{\pm0.000}$ \\

\regress{} &
4.28 & 12.07 &
$0.373_{\pm0.005}$ &
\best{$0.991_{\pm0.000}$} \\

\distill{} &
6.11 & 12.50 &
$0.516_{\pm0.004}$ &
\best{$0.991_{\pm0.001}$} \\

\shuffle{} &
8.41 & 15.51 &
$0.540_{\pm0.023}$ &
$0.958_{\pm0.002}$ \\
\bottomrule
\end{tabular}
\end{table}

\shuffle{} has both the largest raw rollout error ($8.41$, against $4.43$ for \visual{}) and the largest Copy denominator ($15.51$, compared with approximately $12$ for the other methods): the mispaired objective spreads its latents further apart while predicting them less well. Its manipulated-object decodability is also the lowest of any arm ($0.958$). Incorrect visual--physical pairing therefore degrades snapshot content, forecastability and control at once, rather than trading one against another.

\begin{figure}[t]
\centering
\includegraphics[width=\linewidth]{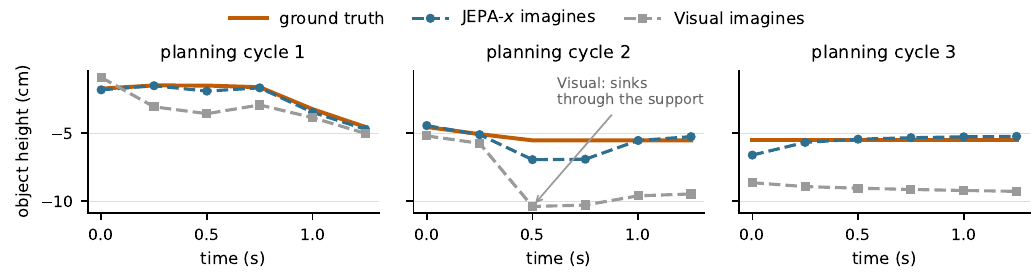}
\caption{\textbf{Under the same actions, \ours{} imagines the object coming to rest on its support.}
Decoded height of the manipulated object during a stacking episode: ground truth against each model's own rollout over the identical expert action chunks, one panel per planning cycle.
\visual{}'s imagined object passes through the support and stays there; \ours{} has a mean terminal error of $1.4$~cm across the three cycles.}
\label{fig:imagination}
\end{figure}

\paragraph{Imagined trajectories under a fixed action sequence.}
\label{app:imagination}
Relative drift summarizes forecastability as a scalar; Figure~\ref{fig:imagination} shows what it looks like on one trajectory in physical units.
We take a stacking episode and, at the start of each planning cycle, roll each model's predictor from the same observation over the same recorded expert action chunks.
The manipulated object's position is decoded from the predicted latents by a probe fitted on training episodes only ($R^2=0.984$ and $0.988$ for the two models); first-step decoding error remains below $2$~cm in every cycle. With identical actions and starting observations, the plots compare decoded rollout behavior; residual state-decoder error remains.
Mean terminal error across the three cycles is $1.4$~cm for \ours{} and $3.1$~cm for \visual{}.

\paragraph{Rank-controlled forecastability.}
\label{app:rank}
Relative drift compares a fitted predictor against a temporal-persistence baseline, so a representation of lower intrinsic dimension could in principle be easier for a fixed-capacity probe to fit.
\ours{} does produce a lower-dimensional latent: measured as the participation ratio of the eigenspectrum, its effective rank is $56.5_{\pm0.2}$ against $95.7_{\pm2.4}$ for \visual{}, out of $192$ dimensions, and this holds for every training seed.
To separate dimensionality from content we project each representation onto its leading $k$ principal components and refit the identical probe.
The basis is fitted on the training episodes only and then applied to all episodes, so the held-out split does not enter the projection.
Everything else is held fixed: the same episodes, split, actions, probe architecture, and optimization budget, with three probe seeds per condition.

\begin{table}[h]
\caption{\textbf{Matching effective rank does not close the forecastability gap.}
Relative drift after projection onto the leading $k$ principal components, reported as mean $\pm$ sample SD over three training seeds.}
\label{tab:rank}
\centering
\small
\setlength{\tabcolsep}{6pt}
\begin{tabular}{lcc@{\hskip 10pt}cc}
\toprule
& \multicolumn{2}{c}{Effective rank}
& \multicolumn{2}{c}{Relative drift $\downarrow$} \\
Projection & \ours{} & \visual{} & \ours{} & \visual{} \\
\midrule
None (native) & $56.5$ & $95.7$ & \best{$0.104_{\pm0.004}$} & $0.359_{\pm0.018}$ \\
$k=57$ & $51.7$ & $52.8$ & $0.121_{\pm0.008}$ & $0.453_{\pm0.022}$ \\
$k=32$ & $30.7$ & $30.4$ & $0.175_{\pm0.011}$ & $0.402_{\pm0.031}$ \\
$k=16$ & $15.7$ & $15.6$ & $0.193_{\pm0.011}$ & $0.379_{\pm0.025}$ \\
$k=8$  & $\phantom{0}7.9$ & $\phantom{0}7.9$ & $0.214_{\pm0.016}$ & $0.407_{\pm0.033}$ \\
\bottomrule
\end{tabular}
\end{table}

At $k=57$, chosen to approximate \ours{}'s native effective rank, \visual{} and \ours{} reach measured ranks of $52.8$ and $51.7$, with drift of $0.453$ and $0.121$, respectively.
Under more aggressive projection both arms degrade and the ratio narrows, from $3.5\times$ at native width to $1.9\times$ at rank $7.9$, but the gap never closes.
The retained variance shows why the two arms respond differently: at $k=57$, \ours{} retains $95.1\%$ of its variance while \visual{} retains $67.9\%$, so the same nominal width discards far more of the visual-only representation.
Compression never improves the visual baseline, and \ours{} at $k=8$ remains more forecastable than \visual{} at any tested width. The forecastability gap is therefore not explained by effective rank alone.

\section{Privileged-State Interface}
\label{app:interface}

This section gives the exact dimensions and normalization of the privileged-state interface introduced in Section~\ref{sec:method}. Each object token concatenates a fixed-width canonical-geometry descriptor, bounding-box half-extents, a flattened $3\times3$ rotation matrix, and position. Our experiments use an $8^3$ average-pooled signed-distance field as the geometry descriptor. The effector token carries end-effector rotation, position, and finger opening, while a learned table token completes the set. Scenes are padded to six object slots with unused slots masked from both self-attention and pooling, giving eight tokens in total. All physical quantities are normalized using training-set statistics.

The privileged input describes only the instantaneous physical configuration and excludes velocities, wrenches, contact forces, and goal-relative quantities. Both branches receive length-$H$ histories, so they infer motion from temporal context and action.

\section{Implementation Details}
\label{app:impl}

\paragraph{Encoders.}
The visual encoder is a ViT-Tiny \citep{dosovitskiy2021image} with $192$-dimensional features, patch size $14$, and $224\times224$ input resolution (5.50M parameters). The physical encoder is a three-layer, four-head Transformer of width $192$ with hidden width $256$ (1.55M parameters), consuming the token set defined in Appendix~\ref{app:interface}.

\paragraph{Predictor and objective.}
The action-conditioned predictor is a six-layer, $16$-head Transformer operating on $H=3$ latent frames, with head dimension $64$, feed-forward width $2048$, and dropout $0.1$. Actions enter through zero-initialized AdaLN. Each action chunk spans five environment steps with a one-step prediction offset. The four source--target terms in Equation~\ref{eq:total} receive equal weight. SIGReg is applied separately to the two branches with weight $\beta=0.09$, $17$ knots, and $1024$ projections.

Both future latents are online encoder outputs; the \ours{} prediction path contains no stop-gradient, exponential-moving-average encoder, or separate target network.

\paragraph{Optimization.}
Core matched methods use AdamW with learning rate $7\times10^{-5}$ and weight decay $10^{-3}$, batch size $256$, bfloat16 precision, gradient clipping at $1.0$, and a linear-warmup cosine-annealing learning-rate schedule. All multi-task models are trained for $40$ epochs over the augmented corpus. No image augmentation is applied at training time; input diversity comes from the corpus-level action-noise augmentation described under \emph{Data} below. Training-time auxiliary components differ only where required by the stated objective or ablation.

\paragraph{Planning.}
The multi-task experiments use a horizon of five macro-steps, each containing five environment actions, with $300$ candidates, $30$ CEM refinement iterations, and $30$ elites ($10\%$). The search distribution is initialized at zero mean and unit variance in the latent action space and re-fitted to the elites at each iteration, with the current mean always evaluated as one candidate. The planner executes the entire five-macro-step sequence, that is $25$ environment actions, before replanning from the resulting state; the executed segment is therefore open-loop and planning is repeated once per segment rather than at every macro-step. The unconditional action autoencoder used by $z$-CEM is a variational autoencoder with a $32$-dimensional latent whose encoder and decoder are two-hidden-layer MLPs of width $512$, mapping a flattened $125$-dimensional action chunk to and from the latent. It is trained on the same corpus for $20{,}000$ steps with AdamW at learning rate $10^{-3}$, weight decay $10^{-5}$, and cosine decay, under a reconstruction plus KL objective with KL weight $\beta=0.1$. It receives neither observations nor goals, and the same action prior is used for every method.

Directly sampling raw action sequences can produce temporally incoherent candidates far from the action distribution seen during training. Searching in the autoencoder latent space instead provides temporal structure: each latent candidate decodes to a coherent action chunk, and CEM refines a distribution over latent codes without conditioning candidates on the current task or scene.

Single-task experiments use the released LeWM action-space CEM planner and its original evaluation budget.

\paragraph{Data.}
The multi-task corpus is the union of one clean shard and four action-noise shards, giving $29{,}260$ episodes and approximately $1.44$M transitions across the $22$ configurations. The clean shard contributes $9{,}900$ episodes ($450$ per configuration) collected with noise disabled. The four noise shards contribute $5{,}104$, $4{,}752$, $4{,}752$ and $4{,}752$ episodes and differ only in noise scale.

Noise is injected into the expert controller during collection rather than added to a recorded trajectory, so every noisy episode is a physically consistent rollout of a perturbed policy. It acts at two levels.
At each replanning tick the target waypoint is displaced by $\mathcal{N}(0,\sigma_p^2 I_3)$ and the target yaw by $\mathcal{N}(0,\sigma_{\mathrm{yaw}}^2)$; the displacement is held fixed for the whole replan segment, so replanning cannot average it away, and the perturbed target is then clipped into the reachable workspace.
At every control step an Ornstein--Uhlenbeck process perturbs the action itself,
\begin{equation}
x \leftarrow (1-\theta)\,x + \varepsilon,
\qquad
\varepsilon \sim \mathcal{N}\!\bigl(0, \sigma_{\mathrm{step}}^2\, s^2\bigr),
\qquad
\sigma_{\mathrm{step}} = \sigma_a\sqrt{2\theta-\theta^2},
\end{equation}
with $\theta=0.15$ and per-dimension scale $s=[1,1,1,1,0.25]$, so the gripper channel receives a quarter of the amplitude. The step variance is chosen so the stationary per-dimension standard deviation equals $\sigma_a$. The sample is added to the controller's output and the sum is clipped to $[-1,1]$.

The $\sigma$ values carry a per-family calibration factor $k$, fitted by bisection so the expert's failure rate on that family lands in a target band. The four shards are the resulting tiers: the band shard targets a $20$--$30\%$ failure rate, and the s50, s20 and s0 shards target $45$--$55\%$, $75$--$85\%$ and at least $98\%$, giving $k=0.0625$, $0.0859$, $0.1094$ and $0.325$ with $(\sigma_p,\sigma_a)$ of $(2.4\times10^{-4}, 6.3\times10^{-3})$, $(3.3\times10^{-4}, 8.6\times10^{-3})$, $(4.2\times10^{-4}, 1.1\times10^{-2})$ and $(1.2\times10^{-3}, 3.3\times10^{-2})$. Yaw waypoint noise is disabled throughout ($\sigma_{\mathrm{yaw}}=0$). Noise is therefore graded by how much it degrades the expert, not by a fixed magnitude, and the corpus spans from a lightly perturbed expert to one that essentially never succeeds.

The LeWM datasets are used as released. Code, the corpus-generation pipeline, and evaluation harnesses will be released.

\section{Ablation Details}
\label{app:ablimpl}

All ablations use the same training data, visual backbone, latent dimensionality, optimization hyperparameters, and training schedule as \ours{}. Predictor parameterization and cross-modal prediction terms are changed only where required by the intervention being tested.

\paragraph{\textsc{Cross-only}.}
\textsc{Cross-only} retains all four source--target prediction terms from \ours{} and therefore preserves corresponding cross-prediction, but replaces the shared predictor with separate visual and physical predictors. This removes predictor-parameter sharing while preserving corresponding cross-prediction and direct cross-modal transition coupling.

\paragraph{\textsc{Share-only}.}
\textsc{Share-only} retains the shared action-conditioned predictor but removes the two cross-modal prediction terms. Each modality is therefore trained only to predict its own future representation through the common predictor. This isolates predictor sharing without directly grounding either prediction in the future of the other modality.

\paragraph{\regress{}.}
\regress{} keeps the visual branch and its self-prediction term unchanged and adds a per-frame state-regression head. A two-layer MLP of width $512$ maps each visual latent to an $85$-dimensional target formed by six object slots, each contributing a flattened $3\times3$ rotation matrix and a position, followed by the effector's rotation, position, and finger opening:
\begin{equation}
\mathcal{L}_{\mathrm{reg}}
=
\frac{\sum_{d} m_d \left( \left[r_\omega(z_t^o)\right]_d - s_{t,d} \right)^2}{\sum_{d} m_d},
\end{equation}
where $r_\omega$ is the regression head and $m_d$ masks the dimensions of unoccupied object slots; the effector dimensions are always included. Targets use the same training-set normalization as the privileged branch. The term is added to the visual self-prediction and isotropy losses with weight $1.0$. No physical encoder or cross-modal prediction is present, and the regression head is unused at deployment.

\paragraph{\distill{}.}
\distill{} replaces predictive cross-modal coupling with pointwise regression. Each branch retains its own within-modality prediction term and is advanced by its own predictor, each followed by its own output projection that maps the predictor's hidden state back to the embedding dimension. These projections belong to the prediction path only: they are applied to predictor outputs and play no part in the cross-branch coupling. Both within-modality prediction terms ($m=n$ in Equation~\ref{eq:total}) therefore remain active, and the physical encoder is still trained, by its own prediction term together with the isotropy regularizer. The two modalities are coupled only by aligning the raw encoder outputs, with no projection applied on either side, the visual latent being regressed onto a stop-gradient physical latent:
\begin{equation}
\left\lVert
z_t^o-\operatorname{sg}(z_t^s)
\right\rVert_2^2.
\end{equation}
The physical target remains paired with its observation, so this baseline preserves sample-level correspondence while constraining representation content rather than cross-modal predictive transitions.

\paragraph{\textsc{Align-only}.}
\textsc{Align-only} keeps the shared predictor and both branches' within-modality prediction terms but excludes the two cross-modal terms from the loss, and adds an explicit symmetric alignment
\begin{equation}
\lambda_{\mathrm{align}}\,\bigl\lVert z^o - z^s \bigr\rVert_2^2 ,
\end{equation}
with $\lambda_{\mathrm{align}}=1$. The term is applied to the full encoded window rather than only to the prediction targets, and neither branch is detached. 

\paragraph{\shuffle{}.}
\shuffle{} retains all four \ours{} source--target terms and the shared predictor, but reads the privileged stream from a fixed partner episode:
\begin{equation}
Z_t^{s} \leftarrow Z_t^{s,\pi(e)},
\qquad
z_{t+1}^{s} \leftarrow z_{t+1}^{s,\pi(e)},
\end{equation}
where $e$ is the current episode and $\pi$ is a derangement of the training episodes, drawn once before training and held fixed.
No episode is paired with itself, and each visual trajectory sees the same incorrect partner throughout training, so the mis-pairing is a consistent alternative pairing rather than a fresh random pairing at every step.
Partners are drawn across the whole corpus, so a visual trajectory may be paired with a privileged trajectory from a different configuration; the privileged geometry, slot occupancy, and pair-validity masks follow the partner, and the partner window is taken at the same relative position within the partner episode.
Because context and future are read from the same partner, each privileged sequence remains an intact trajectory that obeys the environment dynamics, and the isotropy regularizer operates on these intact trajectories as the privileged branch's own task.
Pixels and actions stay on the true episode.
The intervention therefore preserves the physical trajectories themselves while replacing their pairing with the visual stream by a fixed, incorrect one.

\section{Predicting the Physical Future in Raw State Space}
\label{app:futaux}

The closest alternative to cross-predictive grounding is to keep the visual
model unchanged and add an action-conditioned head that predicts the physical
future directly. Visual self-prediction and SIGReg are exactly \regress{}'s;
what changes is the privileged supervision, which becomes an MLP mapping
$(z^o_t, a_t)$ to the next privileged state rather than the current one. Only
the target space then separates this arm from \ours{}: the raw physical future
here, the physical branch's representation of that future there. Deployment is
pixel-only, identical to \visual{} and \regress{}; the head is unused.

We train two variants for three seeds each at the same budget. \textsc{Fut-abs}
regresses $s_{t+1}$. \textsc{Fut-delta} regresses $s_{t+1}-s_t$, which is the
sharper test: measured on the corpus in the model's normalized units, $s_t$
already accounts for about $99\%$ of the squared norm of $s_{t+1}$ -- most of
the $85$-dimensional target is rotation, which barely moves across a $250$\,ms
row -- so an absolute head can score well by decoding the present state, which
is what \regress{} already supervises.

\begin{table}[ht]
\caption{Per-family control success (\%) for the two future-state auxiliary
variants, with \visual{} and \regress{} for reference. Mean $\pm$ sample SD
over three training seeds, same protocol as Table~\ref{tab:family-control}.}
\label{tab:futaux}
\centering
\small
\setlength{\tabcolsep}{5pt}
\begin{tabular}{lcccc}
\toprule
Family & \textsc{Fut-delta} & \textsc{Fut-abs} & \visual{} & \regress{} \\
\midrule
Insertion & $54.0_{\pm6.9}$  & $56.7_{\pm22.7}$ & $60.7_{\pm10.1}$ & $54.7_{\pm9.2}$ \\
Grasping  & $39.3_{\pm19.4}$ & $40.0_{\pm6.9}$  & $34.0_{\pm10.0}$ & $48.0_{\pm3.5}$ \\
Stacking  & $60.0_{\pm2.0}$  & $60.0_{\pm5.3}$  & $58.7_{\pm4.2}$  & $64.0_{\pm10.4}$ \\
Placement & $52.0_{\pm4.0}$  & $56.0_{\pm4.0}$  & $54.0_{\pm2.0}$  & $46.0_{\pm8.7}$ \\
Tossing   & $72.7_{\pm4.6}$  & $77.3_{\pm6.4}$  & $79.3_{\pm4.6}$  & $76.7_{\pm4.2}$ \\
Pushing   & $30.0_{\pm3.5}$  & $36.7_{\pm8.1}$  & $34.7_{\pm2.3}$  & $32.7_{\pm2.3}$ \\
\midrule
All six   & $51.3_{\pm4.8}$  & $54.4_{\pm3.1}$  & $53.6_{\pm2.5}$  & $53.7_{\pm4.1}$ \\
\bottomrule
\end{tabular}
\end{table}

Neither variant separates from the visual baseline on any of the three
measurements. Control is $51.3_{\pm4.8}$ for \textsc{Fut-delta} and
$54.4_{\pm3.1}$ for \textsc{Fut-abs}, against $53.6_{\pm2.5}$ for \visual{} and
$53.7_{\pm4.1}$ for \regress{}: the spread across variants is smaller than the
spread across seeds. Relative rollout drift is $0.359_{\pm0.010}$ and
$0.386_{\pm0.005}$, against $0.361_{\pm0.024}$ for \visual{} and $0.104$ for
\ours{}. Manipulated-object position decodability is $0.978_{\pm0.001}$ and
$0.986_{\pm0.000}$; only the absolute variant rises above \visual{}'s $0.978$,
in the direction of \regress{}'s $0.991$, which is what its present-state-heavy
target predicts.

Predicting the physical future is therefore not sufficient on its own, and the
target space is what distinguishes this arm from \ours{}. Supervising the raw
next state leaves forecastability and control at the visual baseline whether the
head is asked for the state itself or for its increment; the gains reported in
Section~\ref{sec:experiments} require predicting the future \emph{in the
physical branch's learned representation}, through the shared predictor.

\section{Qualitative Rollouts}
\label{app:rollouts}

\begin{figure}[p]
\centering
\includegraphics[width=\linewidth]{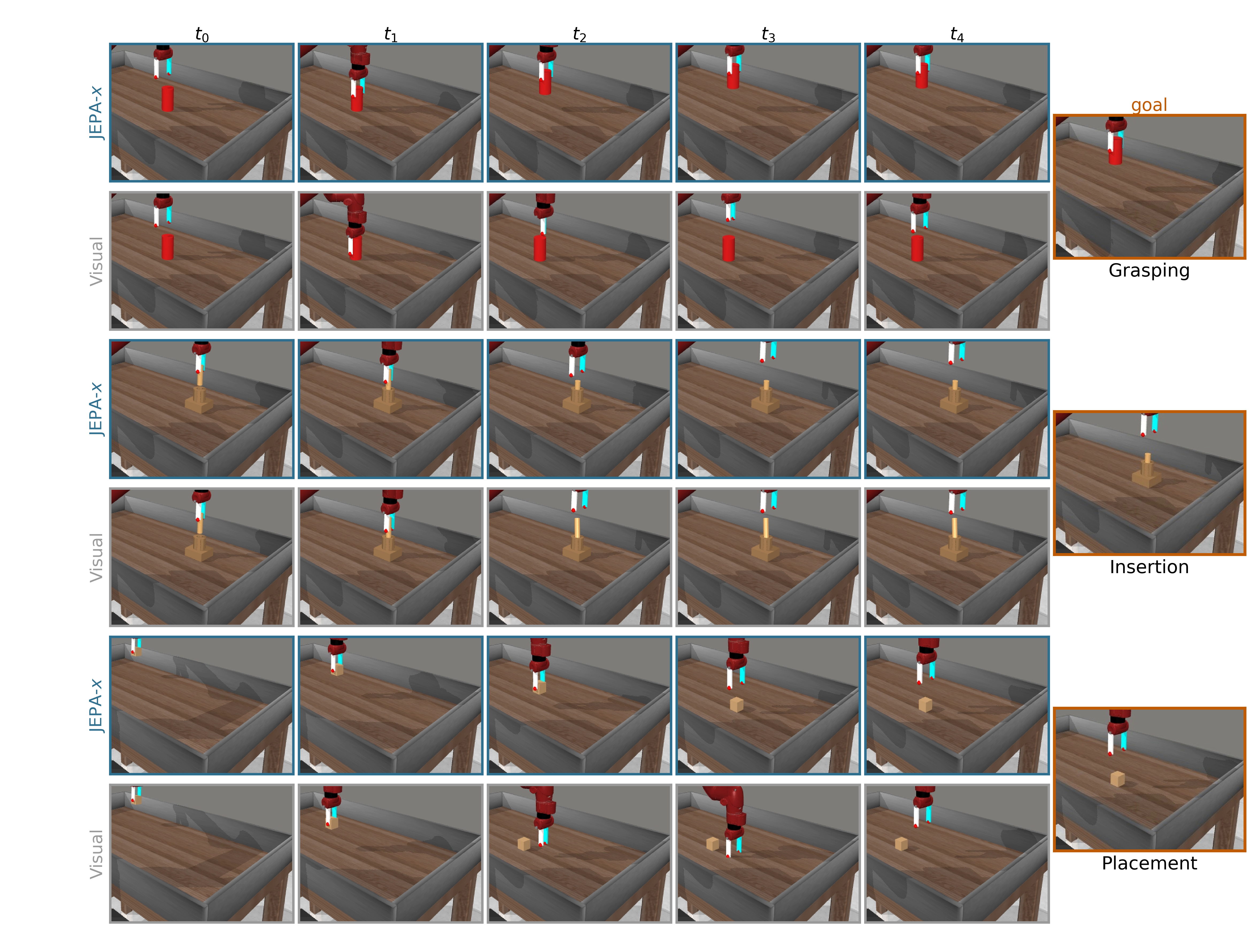}
\caption{\textbf{\ours{} against \visual{} on the same scenario: grasping, insertion, and placement.}
Each family contributes a pair of rows executed from the same initial state using the same planner, \ours{} above and \visual{} below, cropped to the working volume. The goal is shared by both arms and is therefore shown once per comparison, spanning the family's two rows.
Scenarios are the representative episodes in which the two arms disagree, so each pair is a case \ours{} completes and \visual{} does not; per-family success rates are in Table~\ref{tab:family-control}.}
\label{fig:rollouts}
\end{figure}

\begin{figure}[p]
\centering
\includegraphics[width=\linewidth]{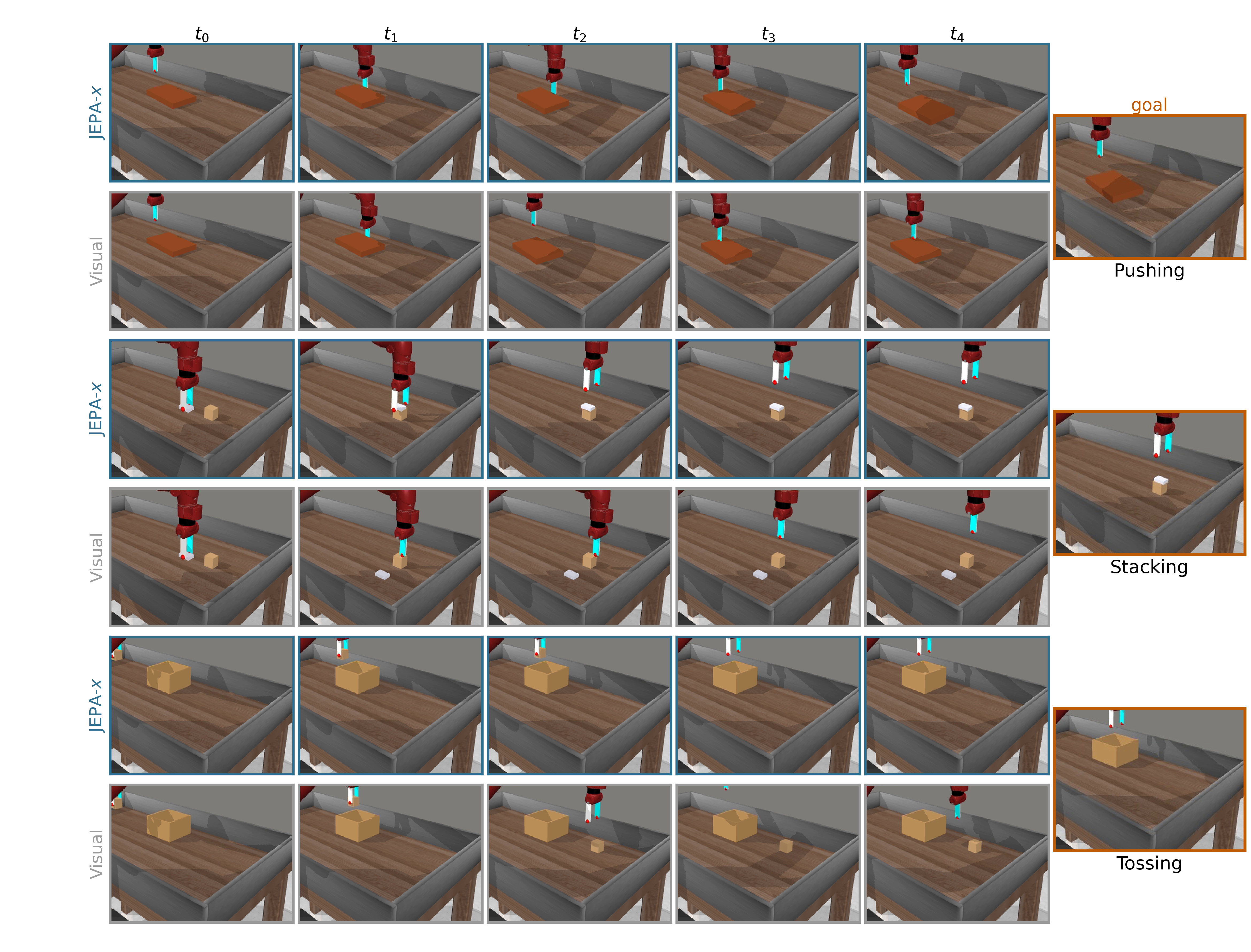}
\caption{\textbf{\ours{} against \visual{} on the same scenario: pushing, stacking, and tossing.}
Same protocol as Figure~\ref{fig:rollouts}: paired rows from one initial state, \ours{} above and \visual{} below, with the shared goal shown once per comparison. In stacking, \visual{} leaves the eraser beside the block; in tossing, it leaves the object outside the carton.}
\label{fig:rollouts2}
\end{figure}

Figures~\ref{fig:rollouts} and~\ref{fig:rollouts2} show paired qualitative comparisons between \ours{} and \visual{} across all six evaluated interaction families. For each family, both methods are executed from the same initial state using the same planner. We show a representative scenario on which the methods disagree, with \ours{} succeeding and \visual{} failing. Frames are cropped to the working volume because the manipulated object occupies only a small fraction of the uncropped camera view, making criteria such as a $5$\,cm lift difficult to inspect at print scale. Quantitative comparisons are reported in Figure~\ref{fig:multitask-control} and Table~\ref{tab:family-control}.

\end{document}